# Significance of the levels of spectral valleys with application to front/back distinction of vowel sounds

T. V. Ananthapadmanabha

Voice and Speech Systems, Bangalore.

*tva.blr@gmail.com*

A. G. Ramakrishnan

MILE Lab, EE Department, IISc, Bangalore

*ramkiag@ee.iisc.ac.in*

Shubham Sharma

MILE Lab, EE Department, IISc, Bangalore

*shubham@mile.ee.iisc.ac.in*

*Anantha, Ramakrishnan and Shubham, JASA*

## Abstract

An objective critical distance (OCD) has been defined as that spacing between adjacent formants, when the level of the valley between them reaches the mean spectral level. The measured OCD lies in the same range (*viz.*, 3 to 3.5 bark) as the critical distance determined by subjective experiments for similar experimental conditions. The level of spectral valley serves a purpose similar to that of the spacing between the formants with an added advantage that it can be measured from the spectral envelope without an explicit knowledge of formant frequencies. Based on the relative spacing of formant frequencies, the level of the spectral valley, $V_I$ (between $F_1$ and $F_2$) is much higher than the level of $V_{II}$ (spectral valley between $F_2$ and $F_3$) for back vowels and vice-versa for front vowels. Classification of vowels into front/back distinction with the difference $(V_I - V_{II})$ as an acoustic feature, tested using TIMIT, NTIMIT, Tamil and Kannada language databases gives, on the average, an accuracy of about 95%, which is comparable to the accuracy (90.6%) obtained using a neural network classifier trained and tested using MFCC as the feature vector for TIMIT database. The acoustic feature $(V_I - V_{II})$ has also been tested for its robustness on the TIMIT database for additive white and babble noise and an accuracy of about 95% has been obtained for SNRs down to 25 dB for both types of noise.

## I. INTRODUCTION

It is well established that there is considerable variability in the formant frequencies measured during the mid-part of a vowel sound spoken in the same context by different speakers of the same gender and same dialect[1]. Native listeners have no difficulty in identifying the phonetic quality of a vowel despite a wide acoustic variability. This finding has created considerable research interest related to vowel identification based both on perceptual and objective criteria. One of these approaches is related to the influence of the spacing between adjacent formants on the perceived vowel quality. Our interest in this paper is related to this approach. It is well recognized that the spacing between two adjacent formants determines the level of spectral valley between them, the level being shallower when the spacing is less and deeper when the spacing is more. Hence, the influence of spacing between two formants can as well be studied indirectly using the level of spectral valley between them. The advantage is that the level of spectral valley can be measured from the spectral envelope without an explicit knowledge of formant data and at the same time not sacrificing the information about the spacing between the formants. In this paper, we explore the significance of the level of spectral valley.

Delattre *et al.* of Haskins Lab reported in 1952[2] an interesting experimental study of vowel quality. They showed that vowels synthesized with a single formant of an appropriately chosen resonant frequency match well with the perceptual quality of back vowels synthesized with two formants, as judged by subjects with a training in phonetics. This equivalence of phonetic quality of a two-formant vowel to that of a single formant vowel is referred to as "spectral integration" in the literature. Surprisingly, this integration did not seem to emerge in the case of front vowels, except for vowel 'i' as an extreme case. Chistovich *et al.*[3] conducted subjective experiments to derive the condition under which the spectral integration occurs. They kept $F_2$ fixed and varied the spacing between $F_1$ and $F_2$. Such a two-formant stimulus was compared with a single-formant stimulus for equivalence in vowel quality. They found that perceptual equivalence occurs when the spacing between the formants lies within the range: $3.1 < \Delta Z_c < 4.0$ bark for an $F_2$ value of 1.8 kHz and in the range $3.3 < \Delta Z_c < 4.3$ bark for the $F_2$ value of 1.4 kHz. Although this critical distance, denoted as $\Delta Z_c$, has a wide range of 3.1 to 4.3 bark and the experiment by Delattre *et al.*[2] demonstrated spectral integration for 'u' with a formant separation of 3.94 bark, this perceptual phenomenon is generally known as 3-bark rule in the literature.

Chistovich *et al.*[3] argued that the frequency of the single-formant equivalent corresponds to the center of gravity of the spectrum. Chistovich and Lublinskaya[4] showed that spectral integration occurs even when there is a large change in formant levels when the spacing between formants is less than the critical distance in the range of $3 - 3.5$ bark. Experiments on matching a four-formant vowel to a two-formant vowel also seem to support the so called 3-bark rule[5;6].

The 3-bark rule has been applied in the context of vowel classification[7], where $(F_1 - F_0)$, $(F_2 - F_1)$ and $(F_3 - F_2)$ differences in bark have been used. The authors[7] argue that these differences compared with a threshold of 3 bark, rather than the actual values of the formant frequencies, is of importance for the discrete vowel classes to emerge. Vowel classification accuracy of $87 - 99\%$ has been reported. A practical scheme of vowel classification based on the separation of formants requires an accurate estimation of formant frequencies, which is quite challenging, especially for high $F_0$ vowels[8]. Hence, some researchers have proposed a spectral template approach instead[9;10;11].

Based on the success of using the formant separation as an acoustic feature, Syrdal and Gopal[7] proposed an auditory model for vowel perception which raises some issues to be resolved. The subjective experimental evidence of 3-bark rule has been established mainly for back vowels. The separation $(F_2 - F_1)$ is not distinctive since it is greater than 3 bark for all front vowels as well as for a number of back vowels. Perceptual tests[12] have shown instances of stimuli with the same separation of $F_0$ and $F_1$ producing a high-low distinction and also instances of stimuli with $F_2$ and $F_3$ separation less than 3 bark producing a vowel quality distinction.

Some researchers consider spectral integration to be a general psycho-acoustic phenomenon, not necessarily restricted to vowels, based on perceptual tests conducted on two-tone complex signals[13] and sinusoids replacing formants[14;15]. Although it is generally believed that spectral integration occurs for back vowels, there may be some exceptions as demonstrated for a specific back vowel of Chinese language[16]. Spectral integration has also been reported in the literature with respect to syllable initial stops[17] and glides[18] but the interest in this paper is restricted only to vowel sounds.

**Motivation for the present work**: The subjective critical distance of $3 - 3.5$ bark has been derived using two-formant unrounded synthetic vowels resembling the quality of 'a' or 'e'. It is not known if the same subjective critical distance is also valid for multiple-formant natural vowels with different bandwidths, formant levels, spectral tilt etc. It is extremely time consuming to conduct subjective experiments to deduce the critical distance for all the experimental conditions. With this in view, we propose an objective critical distance (OCD) that may be measured for any given spectral envelope and investigate if such an OCD serves a similar purpose as that of the subjective critical distance.

**About this work**: In Sec.II, an OCD is defined in terms of the level of spectral valley between two formants. The influence of formant spacing, higher formants, formant levels and fundamental frequency on the measured OCD is studied. In Sec.III, the measured OCD is compared with the subjectively determined critical distance published in the literature. In Sec.IV, the importance of the relative level of spectral valley as an acoustic feature for a front/back classification of vowels is presented.

## II. THE OBJECTIVE CRITICAL DISTANCE

### A. Definition

Rather than subjective, we define an OCD, denoted as $\Delta Y_c$. Since the spacing between two formants determines the level of spectral valley between them, we define the OCD in terms of the level of spectral valley as follows: Let us denote the ratio of mean spectral level to the level of the spectral valley between formants $F_1$ and $F_2$ (or $F_2$ and $F_3$) as $V_{12}$ (or $V_{23}$). Then the OCD is that value of the separation between $F_1$ and $F_2$ (or $F_2$ and $F_3$) when $V_{12}$ (or $V_{23}$) becomes 1. For formant spacing less than $\Delta Y_c$, $V_{12}$ (or $V_{23}$) will be less than 1 (the level of the valley is above the mean spectral level) and $V_{12}$ or $V_{23}$ in dB is less than 0. For formant spacing greater than $\Delta Y_c$, $V_{12}$ (or $V_{23}$) will be greater than 1 (the level of the valley is below the mean spectral level) and $V_{12}$ or $V_{23}$ in dB is greater than 0. In the following sub-sections, we study the relation between the relative level of spectral valley (abbreviated as RLSV) and formant spacing using synthetic as well as natural vowels to measure $\Delta Y_c$ and to identify the factors that influence its value.

### B. Influence of formant spacing on $\Delta Y_c$ in a two-formant vowel

A two-formant vowel is synthesized using a second order digital resonator with $F_2$ and $B_2$ kept constant at 1400 and 200 Hz, respectively. The first formant $F_1$ is varied in frequency from 650 to 950 Hz in steps of 50 Hz and its bandwidth is kept constant at 100 Hz. The reason for this choice of parameters is presented later in Sec.III. The magnitude squared spectrum of the impulse response of the cascaded two-formant model is computed. Figure 1 shows the log-magnitude spectra for three choices of formant spacing. The peaks are clearly resolved in the log-magnitude spectra for all the three conditions. However, when the formant spacing is greater than the OCD, the log-spectrum crosses the mean level twice for each formant, once on either side of the formant peak; else when formant spacing is less than the OCD, there are only two, instead of four crossings, for the two formants. Figure 2 shows the RLSV, $V_{12}$, as a function of formant separation in bark. It is not surprising that $V_{12}$ depends on the formant spacing. However, what is worth noting is that $V_{12}$ is nearly zero when the spacing between two formants is near 3 bark. The variable $V_{12}$ (in dB) is positive when formants are wide apart; It is nearly equal to 0 when the spacing approaches about 3.2 bark and negative when the spacing between the formants reduces further.

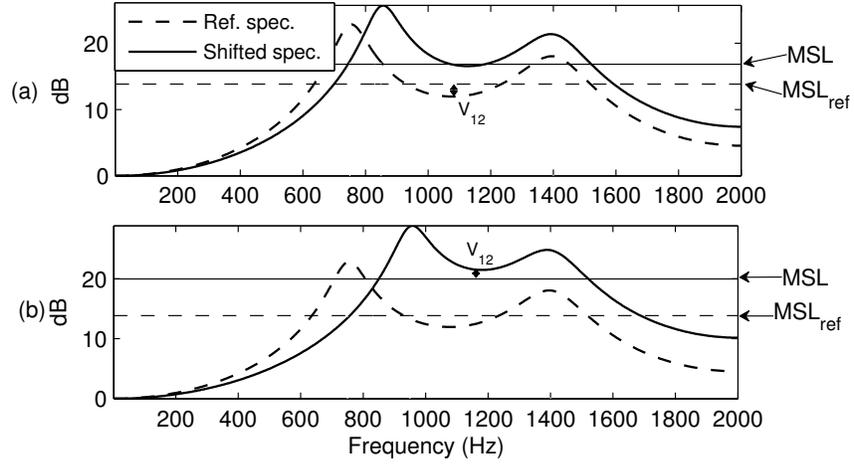

Figure 1: Log-magnitude spectra of synthetic two-formant vowels illustrating the influence of formant spacing on the relative level of spectral valley. Dashed plots: $F_{1_{ref}} = 750$ Hz, $F_{2_{ref}} = 1400$ Hz, $(F_{2_{ref}} - F_{1_{ref}}) = 3.9$ bark, $V_{12} > 1$. (a) Solid plot: $F_1 = 850$ Hz, $(F_2 - F_1) = 3.2$ bark, $V_{12} = 1$ and hence, this $F_1$, $F_2$ separation corresponds to the OCD. (b) Solid plot: $F_1 = 950$ Hz, $(F_2 - F_1) = 2.5$ bark. For these values, $V_{12} < 1$ or its dB value is -ve. (MSL: Mean spectral level)

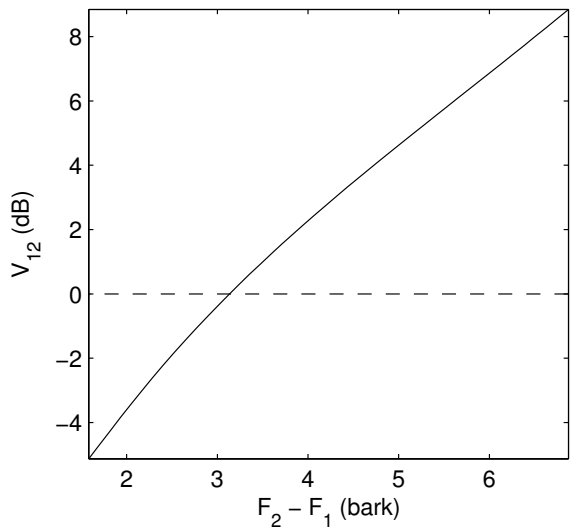

Figure 2: Influence of formant spacing on the relative level of spectral valley.

## C. Influence of higher formants on $\Delta Y_c$

We consider a four-formant vowel with the formant frequencies of a uniform tube at 500, 1500, 2500 and 3500 Hz. As will be discussed in Sec.II.D.1, the choice of bandwidth does not significantly influence $\Delta Y_c$. Hence, for simplicity, bandwidths for all the four formants are kept fixed at 100 Hz. Any other choice for bandwidths could have been made without significantly affecting the results. Also, it is not necessary that the bandwidths of all the formants be equal. In order to control the spacing between $F_1$ and $F_2$, $F_1$ is increased and simultaneously $F_2$ is decreased in steps of 25 Hz, which implies that the vowel quality moves towards that of a back vowel. The log-magnitude spectra are shown in Fig.3 for three selected

 Significance of levels of spectral valleys cases. We note that $\Delta Y_c$ is about 3.6 bark. Thus, even in the presence of higher formants, the OCD between $F_1$ and $F_2$ can be measured unambiguously.

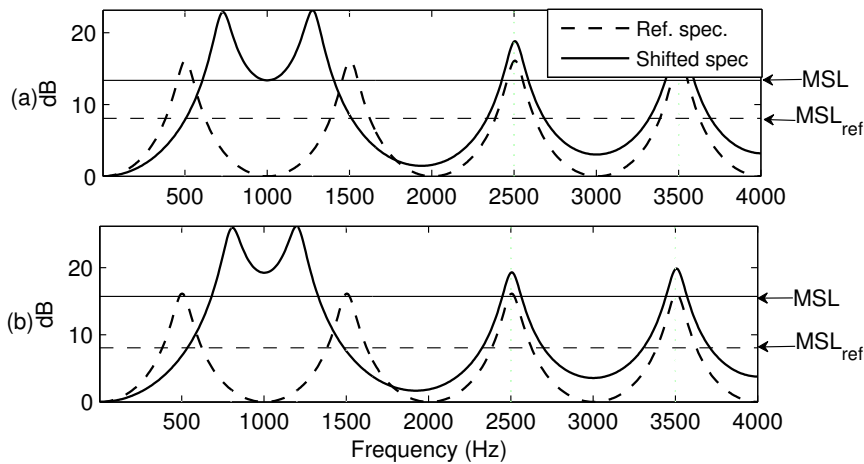

Figure 3: Log-magnitude spectra of synthetic four-formant vowels illustrating the influence of formant spacing on the relative level of spectral valley. Dashed plots: $F_{1_{ref}} = 500$ Hz, $F_{2_{ref}} = 1500$ Hz, $F_{3_{ref}} = 2500$ Hz, $F_{4_{ref}} = 3500$ Hz. $(F_{2_{ref}} - F_{1_{ref}}) = 6.5$ bark. (a) Solid plot: $F_1 = 725$ Hz, $F_2 = 1275$ Hz $(F_2 - F_1) = 3.6$ bark. (b) Solid plot: $F_1 = 800$ Hz, $F_2 = 1200$ Hz $(F_2 - F_1) = 2.6$ bark. (MSL: Mean spectral level)

## D.  Influence of formant levels and $F_0$ on $\Delta Y_c$

The effects of changes in the bandwidth and fundamental frequency are studied for two different cases of formant spacing: Case (a): $F_1$=400 Hz, $F_2$=700 Hz, formant spacing = 2.5 bark. Case (b): $F_1$=600 Hz, $F_2$=1300 Hz, formant spacing = 4.65 bark. The reason for the above choices of $F_1$ and $F_2$ is presented in Sec.III. For synthesis, four-formant model with $F_3 = 2500$ Hz, $F_4 = 3500$ Hz and a sampling frequency of 8000 Hz has been used.

## 1.    Influence of formant levels

Since we are using a cascaded formant model, we can obtain different formant levels by controlling the bandwidths. The bandwidths ($B_1$ and $B_2$) of the first two formants are varied over a wide range to obtain different spectral levels in dB, denoted as $L_1$ and $L_2$, respectively. $B_3$ and $B_4$ are kept fixed at 100 Hz. The magnitude squared spectrum of the impulse response of the cascaded four-formant model is computed. Figure 4 shows RLSV, $V_{12}$ as a function of the difference in formant levels in dB. It is seen that $V_{12}$ in dB is consistently negative for case (a), for x-axis range of about $\pm 6$ dB whereas it is consistently positive for case (b), for x-axis range of about $\pm 10$ dB. Though the difference ($L_1 - L_2$) varies over a wide range of 12 to 20 dB, $V_{12}$ varies only by about 2 dB. Thus a large change in bandwidths or in the formant levels does not significantly influence $V_{12}$ and hence the OCD.

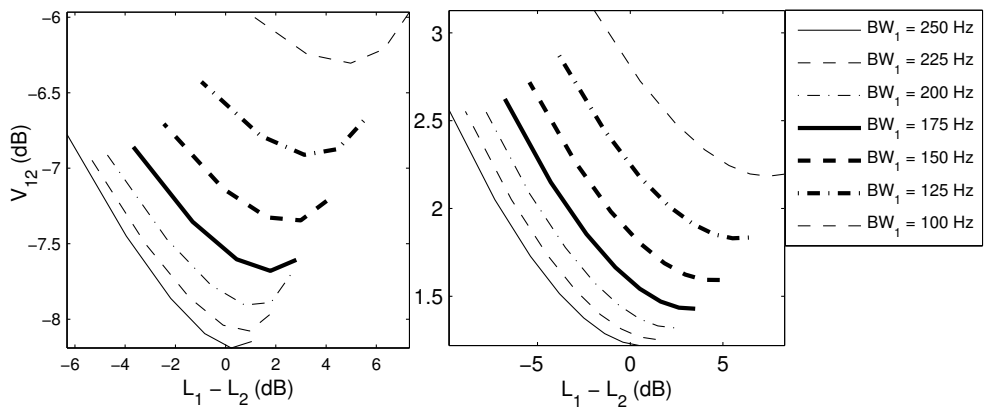

Figure 4: Influence of formant levels on the relative level of spectral valley. $BW_2$ for each $BW_1$ is varied over a wide range. (a) $(F_1, F_2)$ separation < 3 bark and (b) $(F_1, F_2)$ separation > 3 bark.

## 2.  Influence of $F_0$ on $\Delta Y_c$

The response of the 4-formant model is computed for the input of a periodic sequence of impulses for various choices of fundamental frequency $F_0$. Bandwidths of all the formants are kept constant at 100 Hz. Smoothed spectral envelope is obtained using linear prediction of order 8 to measure the level of the spectral valleys, $V_{12_{F_0}}$. Also, the RLSV is measured for the impulse response, which serves as a reference, $V_{12_{ref}}$. The influence of $F_0$ on the difference $(V_{12_{ref}} - V_{12_{F_0}})$ is within $\pm 1$ dB for case (a) and is consistently positive for case (b) as shown in Table 1. Hence, $F_0$ does not have a significant influence on the OCD $\Delta Y_c$.

### E. Objective critical distance, $\Delta Y_c$ for natural vowels

In order to study $\Delta Y_c$ for different vowels, we use the mean formant data of natural vowels of American English published by Peterson and Barney[1]. This is abbreviated as P&B data in this paper. We have synthesized four-formant vowels using the mean formant data of male speakers with a bandwidth of 100 Hz for all the formants and a sampling frequency of 8000 Hz. The fourth formant is fixed at 3500 Hz. Similarly, we have synthesized four-formant vowels using the mean formant data of female speakers with a bandwidth of 100 Hz for all the formants and a sampling frequency of 10000 Hz. The fourth formant is fixed at 4200 Hz for female speakers. We begin with the mean formant data of a vowel and then vary the formant spacing between $F_1$ and $F_2$ by increasing (decreasing) $F_1$ while simultaneously decreasing (increasing) $F_2$ to the same extent. Similarly, the spacing between $F_2$ and $F_3$ is varied. The log-magnitude spectrum is computed from the impulse response of the four-formant vowel.

The measured OCD values are listed in Table 2 in bark for nine vowels. We have considered $V_{23}$ for front vowels instead of $V_{12}$, since for front vowels, the formant spacing between $F_1$ and $F_2$ is very large and $V_{12}$ can be expected to be greater than 1 very often. Similarly, for back vowels $V_{23}$ can be expected to be greater than 1 very often. It is seen that the OCD for ($F_1$, $F_2$) separation for back vowels is in the range of 3.6 to 4.9 bark and for ($F_2$, $F_3$) separation for front vowels is in the range of 1 to 1.8 bark. While the values of the OCD are comparable for male and female speakers, a strong vowel dependency is seen in its behaviour.

 Significance of levels of spectral valleys

## III. COMPARISON OF OCD AND SUBJECTIVE CRITICAL DISTANCES

**Influence of formant spacing**: In Sec.II.B, formant data similar to those used by Chistovich *et al.*[3] have been used. In their study, two parallel bandpass filters were used, with $F_2$ kept fixed and varying $F_1$. The level of $F_2$ was lower than that of $F_1$ and hence we have used $B_2 = 200$ Hz and $B_1 = 100$ Hz. They observed that spectral integration takes place when the spacing between the formants is less than an average critical distance of about 3 bark. The measured OCD based on spectral valley criterion, $\Delta Y_c$, for similar experimental conditions is about 3.2 bark, which matches well with the reported average $\Delta Z_c$.

**Influence of formant levels**: In Sec.II.D.1, we have used formant data and levels similar to those used by Chistovich and Lublinskaya[4]. In their study, two parallel bandpass filters were used and the gains ($A_1$, $A_2$) of the formants were altered. For a given $A_2/A_1$, subjects varied the resonant frequency, $F^*$ of a single formant stimulus so as to match vowel quality to that of a two-formant stimulus. The matched $F^*$ lies between $F_1$ and $F_2$ when the spacing between the formants is less than or equal to the critical distance over a wide range of $A_2/A_1$. Thus, spectral integration is shown to be applicable over a wide range of formant levels. When the formant spacing is greater than $\Delta Z_c$, the responses of the two subjects participating in the experiments were not consistent.

In Sec.II.D.1, we have noted that the RLSV is consistently negative over a wide range of formant levels when the spacing between formants is less than or equal to 3 bark. When the spacing is greater than 3 bark, the RLSV is consistently positive. The objective experiment reported in Sec.II.D.1 gives results similar to those reported in the above subjective study[4].

**Vowel dependency of** $\Delta Y_c$: In Sec.II.E, we considered the formant data of natural vowels and showed that the measured OCD is strongly vowel dependent. We compare this finding with the subjective critical distance reported in the literature.

**Back-vowels,** $(F_1, F_2)$ **Separation**: A change of formant spacing for a uniform tube shifts its quality towards a back vowel. The measured $\Delta Y_c$ between $F_1$ and $F_2$ for a uniform tube is about 3.6 bark, which matches well with the reported $\Delta Z_c$ equal to 3.5 bark[4;13]. The OCD, $\Delta Y_c$ for rounded back vowels lies in the range of 3.9 to 4.9 bark, which is much higher than the subjective critical distance of $3 - 3.5$ bark. The measured high values for the OCD for rounded back vowels get support from other studies: (i) In the original seminal work of Delattre *et al.*[2], it has been shown that spectral integration does take place for vowel 'u' with $F_1 = 250$ Hz and $F_2 = 700$ Hz, having a separation of 3.94 bark, which is greater than the reported subjective critical distance of $3 - 3.5$ bark. (ii) Spectral integration is known to occur for back vowels. In the carefully measured formant data[7], the formant spacing between $F_1$ and $F_2$ for back vowels, for male and female speakers, actually lies in the range of 3.8 to 5.0 bark, which exceeds the subjective critical distance of $3 - 3.5$ bark. (iii) In a subjective experiment on a series of rounded back vowel stimuli[12], $(F_1, F_2)$ separation in the range of 4.2 to 4.7 bark has been used. These evidences, along with the high value of the measured OCD, motivate one to inquire if the subjectively derived critical distance of $3-3.5$ bark can be universally applied for all the experimental conditions.

**Front-vowels, $(F_2, F_3)$ Separation**: To our knowledge, there is no reported study on determining the subjective critical distance for front vowels based on $(F_2, F_3)$ separation, though there have been some studies on equivalent $F_2$ determination of vowels[20;19]. However, in an experiment on front vowel series[12], 'hid' vs 'head', spectral integration is shown to occur, for $(F_2, F_3)$ separation in the range of 1.2 to 1.8 bark, which compares well with the OCD derived for front vowels, shown in Table 2.

## IV. FRONT/BACK CLASSIFICATION USING THE LEVEL OF SPECTRAL VALLEYS

### A. Practical relevance of spectral valleys

The RLSV indicates the relative formant separation with respect to the OCD, i.e., whether the separation is lesser or greater than $\Delta Y_c$ and is not intended to provide the actual value of formant separation. Our interest here is to directly make use of the information related to spectral valley as an acoustic feature for front/back distinction, motivated by the expectation that spectral valley information may be obtained from the spectral envelope without an explicit knowledge of formant data and it subsumes the influence of spectral tilt due to voice source, influence of higher formants and bandwidths.

The relative level of the first (second) spectral valley is denoted as $V_I$ ($V_{II}$). For both back and front vowels, the first (second) spectral valley happens to lie between $F_1$ and $F_2$ ($F_2$ and $F_3$). But we have denoted it as $V_I$ ($V_{II}$) instead of $V_{12}$ ($V_{23}$) since the latter may imply that a knowledge of formant data has been used. For back vowels, level of the first valley is expected to be above the mean level, whereas the second valley is expected to be below the mean level. For front vowels, the trend is the opposite. Hence, for back vowels, $V_I$ is expected to be greater than $V_{II}$ and for front vowels, lesser than $V_{II}$ due to the relative spacing between the formants. Hence, the difference ($V_I - V_{II}$) can also be used as an acoustic feature for front/back distinction. Thus, there are three choices for the feature for front/back distinction, $V_I$, $V_{II}$ and the difference ($V_I - V_{II}$).

## B. Experiments using Peterson and Barney Formant data

We want to investigate if the relative level of a spectral valley serves a similar purpose as that of the formant spacing. Syrdal and Gopal[7] have shown that ($F_3$ - $F_2$) in bark distinguishes front ($< 3$ bark) from back vowels ($> 3$ bark). Normalized histograms of ($F_3$ - $F_2$) in bark for male and female speakers for these two groups of vowels (except 'ar') are shown in Fig.5(a). The accuracy of front/back classification using the 3-bark rule is 99.2%.

P&B[1] provide us with data on the frequencies of the first three formants and only the mean formant levels. In order to derive the bandwidths, we have synthesized vowels using a minus 6 dB per octave source pulse and the mean formant data of male speakers. The bandwidths are adjusted such that the measured formant levels nearly match the mean spectral levels $L_1$, $L_2$ and $L_3$ of the three formants in the published data. Subsequently, these bandwidth values have been used for synthesis. A sampling frequency of 10000 Hz has been used for male and female speakers for the experiments discussed in this section. For male speakers, $F_4 = 3500$ Hz and $F_5 = 4500$ Hz have been used. For female speakers, $F_4 = 4200$ Hz has been used. Formant data of all the available adult male and female speakers have been used [21].

The normalized histograms of $V_{II}$ for the two groups of vowels are shown in Fig.5 (b). A strong similarity is seen between the histograms shown in Figs.5(a) and 5(b), suggesting that $V_{II}$ serves a similar purpose as that of ($F_3$ - $F_2$) in bark for front/back classification of vowels. Front/back classification accuracy of 97.6% is obtained for the entire data using the RLSV $V_{II}$. However, the results reported by Syrdal and Gopal[7] are based on a subset of the measured data corresponding only to perceptually unambiguous vowels.

A similar experiment has been conducted using the relative level of the first spectral valley, $V_I$, which lies between $F_1$ and $F_2$. An accuracy of 98.4% has been obtained for front/back distinction using $V_I$. On the other hand, the separation between $F_1$ and $F_2$ in bark is not effective for front/back distinction. Using the difference ($V_I - V_{II}$) as an acoustic feature for front/back distinction gives an accuracy of 99.01%.

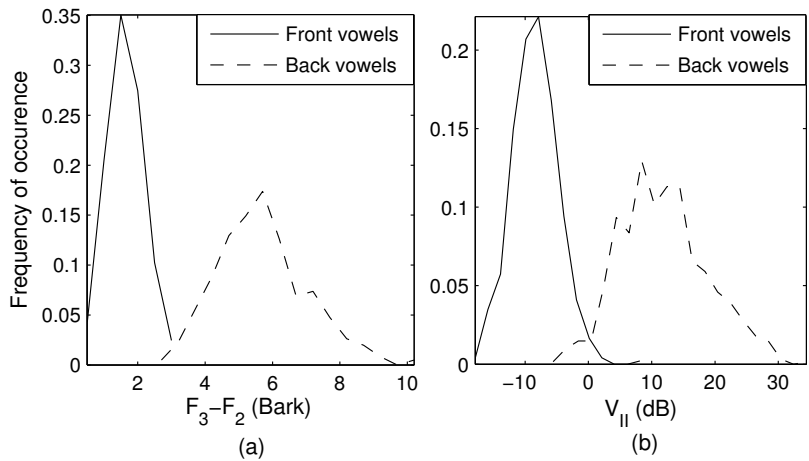

Figure 5: Similarity between features for distinguishing front from back vowels in adult (male and female) speakers. Normalized histograms of (a) $(F_3 - F_2)$ in bark and (b) RLSV $(V_{II})$ in dB.

## C. Experiments using speech databases

The P&B data were obtained for a constrained hVd context and the measurements were taken from a representative interval of each vowel. However, in continuous speech, a strong influence of the context can be expected. In order to study the influence of the context on front/back distinction, experimental results are presented in this subsection on various databases using the acoustic feature $(V_I - V_{II})$. The features $V_I$ and $V_{II}$ are not used as they are dependent on the mean level. However, it may be noted that some monophthong vowels, (especially, 'ux' and 'ax') have been phonetically considered as 'central' instead of being classified as front or back. This aspect will be covered during the following discussion.

## 1.    Experiments on the TIMIT test set database

The TIMIT[22] database is labeled at the phone level and consists of a total of 6300 utterances spoken by 630 speakers belonging to several dialects of North America. The database is divided into the Training and Test sets of 8 dialects, comprising 4620 utterances and 1680 utterances, respectively.

Vowels 'iy', 'ih', 'eh', 'ae', 'aa', 'ah', 'ao', 'uh', 'uw', 'ux' and 'ax' have been used for the experiment. Out of these, the vowels 'iy', 'ih', 'eh' and 'ae' are front vowels and the rest are back vowels. Vowels in all the contexts except those preceded or followed by a nasal or 'r' sound or an aspirated 'h' sound are considered. Reduced vowels of extremely short duration ($< 45$ ms) have been excluded as they tend to be neutralized.

A frame duration of 20 ms has been used with 50 percent overlap between two successive frames. Linear prediction analysis of order 18 is performed on pre-emphasized and windowed frames of vowel segments. From the log spectral envelope obtained from the LPCs, the levels of first and second valleys are measured for each frame over the entire duration of the selected vowel segment. When the mean difference ($V_I - V_{II}$) is greater than a threshold of 5 dB, the vowel is classified as a back vowel else it is classified as a front vowel. The threshold of 5 dB arises due to an intrinsic spectral slope. This intrinsic spectral slope arises due to varying bandwidths and relative spacings of the formants. Spectral slope is also influenced by the recording conditions. Only in the case of a neutral vowel with all the bandwidths being equal, the formant levels as well as the levels at the spectral valleys will be the same. The manual labelling available is then made use of to verify the accuracy of front/back classification. Also, formant data are extracted by solving the roots of the LPC polynomial. Front/back classification has also been done using $F_2$, $F_3$ and $F_1$, $F_2$ spacings in bark. Here, Test set of TIMIT database has been considered and the total number of front and back vowels obtained are given in Table 3.

The normalized histograms of the mean difference $(V_I - V_{II})$ are shown in Fig.6(a) for the back and front vowels. The histogram for back vowels shows a tri-modal distribution. The overlapping region between the histograms of the front and back vowels mostly corresponds to the 'central' vowels, 'ux' and 'ax'. In order to discriminate a central vowel from a front vowel, some other acoustic feature is required. The histogram of back vowels excluding 'ux' and 'ax' shows a clear separation from the histogram of front vowels as shown in Fig.6(b). The cross-over of the histograms occurs around 5 dB. Results reported below exclude the central vowels.

A classification accuracy of 95.7% is obtained for the front vowels. For back vowels, the accuracy is 99.6%, giving an overall accuracy of 96.9%. (See Table 4). The high accuracy has been obtained despite variability in context and speaker characteristics. Using $F_2$, $F_3$ separation, the overall accuracy is 88.9% which is lower than that obtained using valley information. Accuracy using $F_1$, $F_2$ separation is extremely poor, namely, 0.4% for back vowels showing a strong influence of the context or probable errors in the estimation of formant data. Even in the case of P&B data with constrained context, the accuracy is only about 11.3 and 14.6% for 'aa' and 'ao', respectively[7].

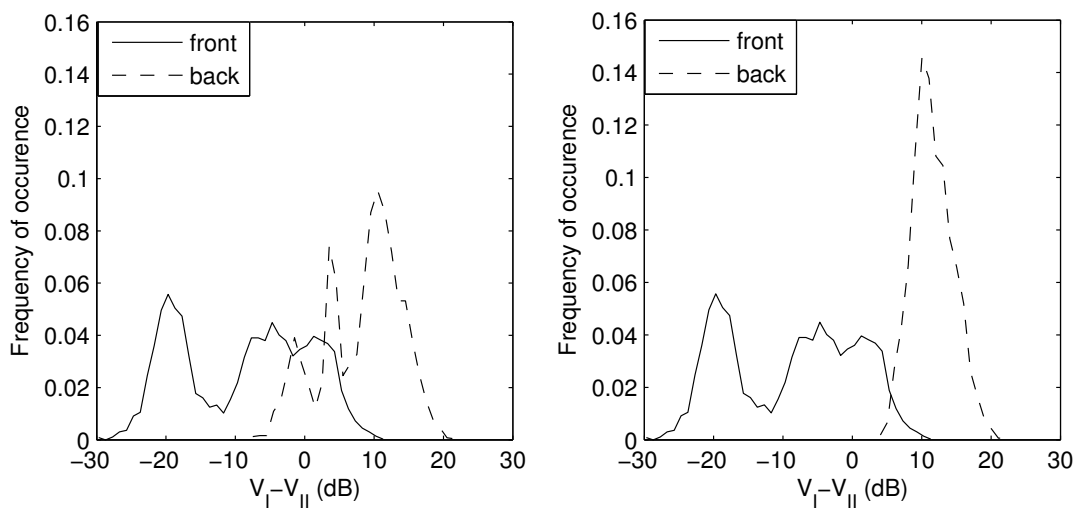

Figure 6: Normalized histograms of front and back vowels of TIMIT database, with back vowels (a) including central vowels (b) excluding central vowels.

## 2.    Experiments on the NITIMIT test set database

To study the performance against channel degradation, NTIMIT test database[23], which is the telephone quality version of the TIMIT, has been used. The utterances in NTIMIT differ from those in TIMIT in two important respects namely, a reduction of bandwidth and a degradation in SNR.

A similar experiment as above is repeated for the NTIMIT database. Using the spectral valley information, a high overall classification accuracy of 89.0% has been obtained despite channel degradation. On the other hand, using $F_2$, $F_3$ separation in bark, the overall accuracy is 72.4%. (See Table 4).

## 3.  Experiments on the MILE Kannada and Tamil Language databases

To test the scalability of the algorithm, we consider databases of two Dravidian languages, Kannada and Tamil. These were recorded in a studio environment for the purpose of the development of text-to-speech synthesis systems at MILE lab, Indian Institute of Science [24]. Each database comprises utterances of phonetically rich sentences spoken by one male speaker and annotated manually at the phone level.

Kannada and Tamil languages have ten vowels, *viz.*, 'i', 'I', 'e', 'E', 'a', 'A', 'u', 'U', 'o' and 'O' of which 'i', 'I', 'e' and 'E' are front vowels and the rest are back vowels. The results are shown in Table 4. High accuracies of 99.9% and 98.4% for Kannada and Tamil, respectively have been obtained using valley information despite a different recording set up and a language other than American English, illustrating the scalability. However, using separation between $F_2$ and $F_3$ provides accuracies of 97.8% and 65.3% for Kannada and Tamil, respectively. Low accuracy for Tamil might have arisen due to error in formant data extraction or due to non-applicability of the 3-bark rule.

## 4.    Experiments on the robustness using the TIMIT test set database

It may appear to be contra-intuitive to use spectral valley information since they represent relatively lower SNR regions in the short-time spectrum. In order to test the robustness of the approach, we have evaluated the performance for additive white and babble noise using the TIMIT test set database. The results are shown in Table 5 for both white and babble noise using both the methods. Using the valley information, the performance is above 95% till 25 dB SNR for white noise and above 96% till 20 dB SNR for babble noise and subsequently deteriorates abruptly, especially for back vowels. This arises since we are using the LP analysis for obtaining the spectral envelope. More robust methods for obtaining the spectral envelope may give better results. Using formant spacing, the overall accuracy is in the range 89 to 91% for white noise and 89 to 90% for babble noise.

## 5.    Comparison with a benchmark

In order to compare the above findings with a benchmark, we have trained a two-layer feed-forward neural network using the Training set of TIMIT database using twelve MFCCs for front/back distinction with the help of MATLAB neural network toolbox[25]. Analysis conditions are the same as used in previous experiments. After training, the neural network is tested using the TIMIT test set. Using ten neurons in the hidden layer, an accuracy of 90.6% has been obtained, which is about 5% lower than that obtained using a simple scalar measure $(V_I - V_{II})$ and about the same as that obtained using the formant spacing. Table 6 also compares the accuracy using the 3-dimensional feature vector consisting of $[V_I, V_{II}, (V_I - V_{II})]$ with that of the threshold classifier using the scalar feature, $V_I - V_{II}$. The classification performance by the trained neural network in both cases is worse than the threshold classification using $V_I - V_{II}$. Therefore, a simple threshold classifier works as good as a trained classifier, while being computationally very simple.

King and Taylor[26] trained a neural network using a 39-dimensional vector for the distinctive feature front/back. They report a frame-wise accuracy of about 86%. They have defined front/back distinction for consonants also. Lee and Choi[27] classified a subset of vowels of TIMIT database into six groups based on tongue advancement using $F_0$ and formants as features and Mahalanobis distance for classification. They report an accuracy of 64.4%.

## V. CONCLUSION

Based on the fact that the level of spectral valley is determined by the formant spacing, a term called objective critical distance has been defined as that spacing when the level of the valley is equal to the mean spectral level. This concept subsumes the formant separation criterion[2;3;7;12;13] and spectral template criterion[9;10;11]. We have shown that the behaviour of the OCD is similar to the subjectively derived critical distance for similar experimental conditions such as different formant spacing, formant levels and presence of higher formants. However, the measured OCD is strongly dependent on the vowel.

It is shown that the relative levels of the spectral valleys, which can be measured from the spectral envelope of a vowel without explicitly estimating the formant values, can be used as an acoustic feature for front/back classification with an overall accuracy of about 95% and is robust for channel degradation and additive noise. We consider the present method to be an acoustic-phonetics knowledge based approach since the acoustic feature used is task specific with a simple threshold based logic, instead of a statistical classifier.

Future research involves looking more closely into the role played by the location as well as the level of spectral valleys for deducing other phonetic features such as high-low, tense-lax, rounded-unrounded and for normalization of intra and inter-speaker differences hopefully to arrive at an invariant vowel space.

# Figure Captions

Figure 1. Log-magnitude spectra of synthetic two-formant vowels illustrating the influence of formant spacing on the relative level of spectral valley. Dashed plots: $F_{1_{ref}} = 750$ Hz, $F_{2_{ref}} = 1400$ Hz, $(F_{2_{ref}} - F_{1_{ref}}) = 3.9$ bark, $V_{12} > 1$. (a) Solid plot: $F_1 = 850$ Hz, $(F_2 - F_1) = 3.2$ bark, $V_{12} = 1$ and hence, this $F_1$, $F_2$ separation corresponds to the OCD. (b) Solid plot: $F_1 = 950$ Hz, $(F_2 - F_1) = 2.5$ bark. For these values, $V_{12} < 1$ or its dB value is -ve. (MSL: Mean spectral level).

Figure 2. Influence of formant spacing on the relative level of spectral valley.

Figure 3. Log-magnitude spectra of synthetic four-formant vowels illustrating the influence of formant spacing on the relative level of spectral valley. Dashed plots: $F_{1_{ref}} = 500$ Hz, $F_{2_{ref}} = 1500$ Hz, $F_{3_{ref}} = 2500$ Hz, $F_{4_{ref}} = 3500$ Hz. $(F_{2_{ref}} - F_{1_{ref}}) = 6.5$ bark. (a) Solid plot: $F_1 = 725$ Hz, $F_2 = 1275$ Hz $(F_2 - F_1) = 3.6$ bark. (b) Solid plot: $F_1 = 800$ Hz, $F_2 = 1200$ Hz $(F_2 - F_1) = 2.6$ bark. (MSL: Mean spectral level).

Figure 4. Influence of formant levels on the relative level of spectral valley. $BW_2$ for each $BW_1$ is varied over a wide range. (a) $(F_1$, $F_2)$ separation $< 3$ bark and (b) $(F_1$, $F_2)$ separation $> 3$ bark.

Figure 5. Similarity between features for distinguishing front from back vowels in adult (male and female) speakers. Normalized histograms of (a) $(F_3$ - $F_2$ in bark and (b) RLSV $(V_{II})$ in dB.

Figure 6. Normalized histograms of front and back vowels of TIMIT database with back vowels (a) including central vowels (b) excluding central vowels.

Table 1: Insignificant influence of fundamental frequency $F_0$ on RLSV. Difference ($V_{12_{ref}} - V_{12_{F_0}}$) is tabulated for different choices of fundamental frequency.

| $F_0$ (Hz) | 100 | 125 | 150 | 175 | 200 | 225 | 250 |
|---|---|---|---|---|---|---|---|
| $F_2$-$F_1$ < 3 bark | −0.30 | −0.49 | 0.09 | −0.48 | −0.78 | 0.99 | −0.51 |
| $F_2$-$F_1$ > 3 bark | 2.23 | 2.44 | 2.29 | 2.79 | 2.60 | 2.45 | 2.32 |

Table 2: The measured OCD values (in bark) for different vowels of adult (male and female) speakers using formant data given by Peterson and Barney[1]. The symbol '∃' denotes uniform tube.

|  | Based on $F_2$-$F_3$ separation | | | | | Based on $F_1$-$F_2$ separation | | | | | |
|---|---|---|---|---|---|---|---|---|---|---|---|
| Vowel | iy | ih | eh | ae | ∃ | ∃ | aa | ao | uh | uw | ah |
| $\Delta Y_c$ (male) | 1.05 | 1.50 | 1.66 | 1.79 | 1.82 | 3.59 | 3.95 | 4.57 | 4.58 | 4.56 | 4.01 |
| $\Delta Y_c$ (female) | 1.08 | 1.32 | 1.46 | 1.67 | 1.82 | 3.59 | 4.33 | 4.89 | 4.86 | 4.93 | 4.26 |

Table 3: Number of front and back vowels considered for TIMIT/NTIMIT, Kannada and Tamil databases.

| Database | No. of front vowels | No. of back vowels | Total number of vowels |
|----------|---------------------|--------------------|------------------------|
| TIMIT/NTIMIT Test set | 4409 | 2122 | 6531 |
| Kannada | 9895 | 22727 | 32622 |
| Tamil | 5602 | 9317 | 14919 |

Table 4: Accuracy (in %) of front/back classification for different speech databases using spectral valley and formant spacing features.

| Feature | Based on $V_I - V_{II}$ | | | Based on $F_3 - F_2$ in Bark | | |
|---|---|---|---|---|---|---|
| Accuracy of | Front | Back | Overall | Front | Back | Overall |
| TIMIT Test set | 95.7 | 99.6 | 96.9 | 86.3 | 94.6 | 88.9 |
| NTIMIT Test set | 89.9 | 90.1 | 89.0 | 82.8 | 49.3 | 72.4 |
| Kannada | 99.9 | 99.9 | 99.9 | 99.5 | 97.0 | 97.8 |
| Tamil | 98.5 | 98.3 | 98.4 | 93.1 | 49.5 | 65.3 |

Table 5: Effect of noise on the front/back classification performance (in %) for different SNR.

| Noise | SNR in dB | 40 | 35 | 30 | 25 | 20 |
|---|---|---|---|---|---|---|
| AWGN | Based on $V_I - V_{II}$ | 97.0 | 97.2 | 96.7 | 95.5 | 88.9 |
| | Based on $F_3 - F_2$ in bark | 89.5 | 88.7 | 90.3 | 90.6 | 90.9 |
| Babble | Based on $V_I - V_{II}$ | 96.8 | 96.8 | 96.9 | 96.7 | 97.2 |
| | Based on $F_3 - F_2$ in bark | 89.0 | 89.4 | 89.5 | 89.8 | 89.3 |

Table 6: Front/back classification performance (in %) of the threshold classifier using scalar feature $(V_I - V_{II})$ compared with those of neural network trained with MFCC and 3-dimensional feature on TIMIT test set.

| Classifier | Feature | Dimension | Accuracy |
|---|---|---|---|
| Neural network based | MFCC | 12 | 90.6 |
| Neural network based | $[V_I, V_{II}, (V_I - V_{II})]$ | 3 | 89.3 |
| Threshold based | $V_I - V_{II}$ | 1 | 96.9 |